\documentclass[sigconf,nonacm]{acmart}

\renewcommand\footnotetextcopyrightpermission[1]{} 

\usepackage{todonotes}

\usepackage{array}

\usepackage[acronym]{glossaries}
\makeglossaries
\usepackage{acronym}
\newcommand*{\sysname}{\fontfamily{ppl}\selectfont \textit{SIGMUS}}

\AtBeginDocument{%
  }

\begin{document}


\title[{\sysname}: Semantic Integration for Knowledge Graphs in Multimodal Urban Spaces]{{\sysname}: Semantic Integration for Knowledge Graphs in Multimodal Urban Spaces\footnotemark}

\author{Brian Wang}
\affiliation{%
  \institution{University of California, Los Angeles}
  \city{Los Angeles}
  \state{California}
  \country{USA}}
\email{wangbri1@g.ucla.edu}
\setcounter{footnote}{1} 
\author{Mani Srivastava \footnotemark}
\email{mbs@ucla.edu}
\affiliation{%
  \institution{University of California, Los Angeles}
  \city{Los Angeles}
  \state{California}
  \country{USA}}


\renewcommand{\shortauthors}{Wang et al.}

\begin{abstract}
Modern urban spaces are equipped with an increasingly diverse set of sensors, all producing an abundance of multimodal data.  Such multimodal data can be used to identify and reason about important incidents occurring in urban landscapes, such as major emergencies, cultural and social events, as well as natural disasters.  However, such data may be fragmented over several sources and difficult to integrate due to the reliance on human-driven reasoning for identifying relationships between the multimodal data corresponding to an incident, as well as understanding the different components which define an incident. Such relationships and components are critical to identifying the causes of such incidents, as well as producing forecasting the scale and intensity of future incidents as they begin to develop.  In this work, we create {\sysname}, a system for \textbf{S}emantic \textbf{I}ntegration for Knowledge \textbf{G}raphs in \textbf{M}ultimodal \textbf{U}rban \textbf{S}paces.  {\sysname} uses Large Language Models (LLMs) to produce the necessary world knowledge for identifying relationships between incidents occurring in urban spaces and  data from different modalities, allowing us to organize evidence and observations relevant to an incident without relying and human-encoded rules for relating multimodal sensory data with incidents.  This organized knowledge is represented as a knowledge graph, organizing incidents, observations, and much more.  We find that our system is able to produce reasonable connections between 5 different data sources (new article text, CCTV images, air quality, weather, and traffic measurements) and relevant incidents occurring at the same time and location.

\end{abstract}

\maketitle


\section{Introduction} \label{introduction}

\begin{figure}
\centering
\includegraphics[width=9cm]{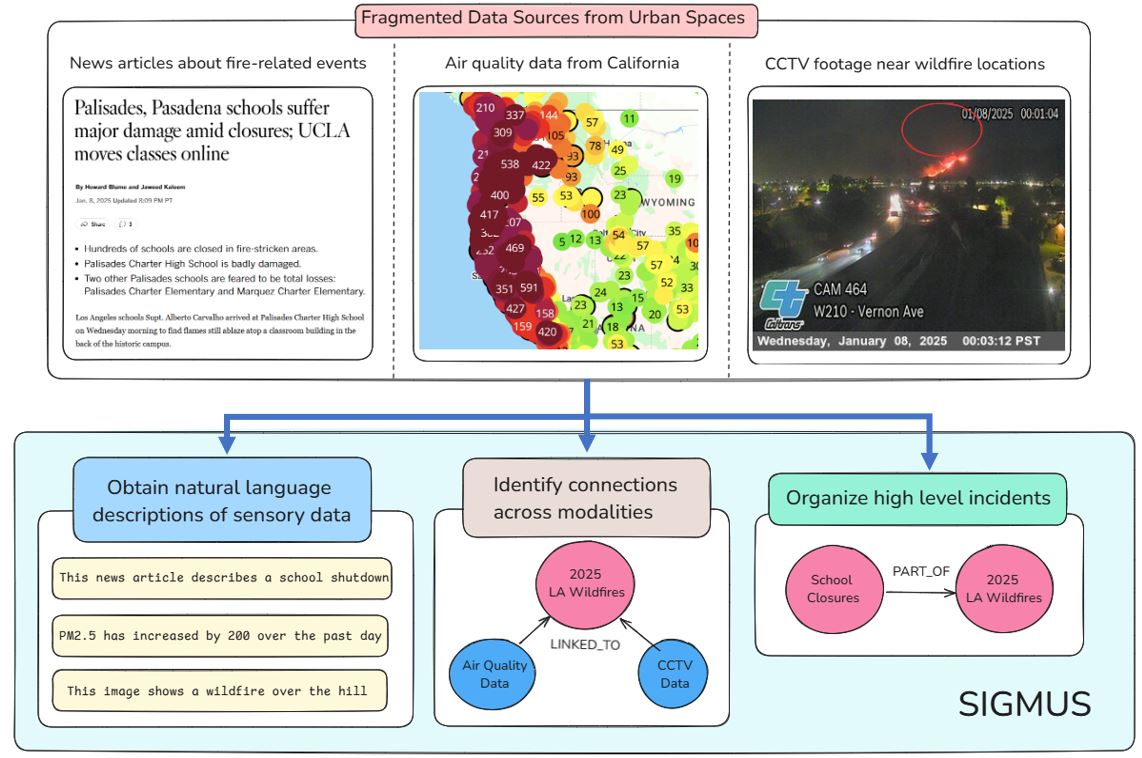}
\caption{High level overview of {\sysname}, which aims to identify incidents in urban spaces and draw connections to sensory data collected in those spaces.}
\label{fig:motivation}
\vspace*{-17px}
\end{figure}

\setcounter{footnote}{2}
\footnotetext{This work was published at the 14th International Workshop on Urban Computing (UrbComp), held at KDD 2025.}

\setcounter{footnote}{3}
\footnotetext{The author holds concurrent appointments as Amazon Scholar and Professor at UCLA, but the work in this paper is not associated with Amazon.}

The proliferation of digital sensors embedded in modern urban infrastructure, coupled with increasingly efficient and robust connectivity technologies has introduced a growing ecosystem of publicly accessible data streams - from traffic cameras to weather stations, environmental monitors, and real-time news feeds.  This ecosystem is further augmented both by human observations made through community-contributed information on social platforms as well as crowd-sourced environmental monitoring through mobile devices.  However, despite the abundance of such rich multimodal data in urban spaces, these data streams often reside in fragmented, siloed systems that lack mechanisms for integrated analysis and reasoning across data streams.  As a result, this ecosystem remains heavily underutilized and lacks tools for understanding and monitoring major phenomena that require analysis across various sources and modalities.

One example of such major phenomena was the catastrophic Palisades Fire (and other concurrent wildfires), which devastated parts of Los Angeles in January 2025.  Understanding the full impact of the wildfires requires an analysis across disparate data sources.  CCTV data, air quality measurements, and news reports each provide a unique and critical insight into how the city may be affected.  Such insight is valuable to emergency service providers, policymakers, health agencies, and future urban planners.  However, this full potential can only be achieved when information across data sources is unified and organized in a human-readable manner.

By unifying disparate data sources, we enable urban knowledge systems to fuse independent reports across different modalities into cohesive, contextualized incidents that can be readily visualized.  This allows for retroactive analysis of important city events.  Furthermore, as many of these data sources publish live data, it becomes possible to create a live knowledge system suited for monitoring and early detection of incidents occurring in urban environments.

Ideally, a live knowledge system for analysis and monitoring of urban environments should describe relationships between data from heterogeneous sources.  Different sources may report data which is related by time, space, as well as the incident itself.  Thus, to overcome the fragmented nature of many data sources, we propose using a knowledge graph to unify and integrate data.

However, many of these relationships within and between data sources may be challenging to infer, as they may be conditioned on the context of the data as well as some degree of world knowledge.  For example, identifying that an air quality trend is related to the report of a house fire depends on both the context (e.g. time and space), but also the physics of the world (e.g. a fire should reduce air quality).  Importantly, such inferences require both world knowledge and a degree of fuzzy reasoning.  

Recently, Large Language Models (LLMs) have been shown to demonstrate significant world knowledge about sensory data \cite{xu_penetrative_2024, yang_drhouse_2024}.  We hypothesize that LLMs, when used to reason about incidents and sensory data, provide valuable insights into the emergence of real-world incidents and their detection through sensory data.  Importantly, the ability to generate meaningful relationships between incidents and different sensory sources (e.g. CCTV, traffic measurements, news articles) requires a significant degree of world knowledge.  In this work, we show the effectiveness of LLMs at the qualitative reasoning required to organize and create relationships between incidents and sensory data.   

The main contribution of this work, {\sysname}, is a system for ingesting data from a variety of data sources (across text, image and tabular data), and integrating multimodal information with identified real-world incidents.  Figure \ref{fig:motivation} showcases the main objectives of our system, allowing for readable information about sensory data and associated incidents to be stored in a knowledge graph.  We provide representative examples to illustrate {\sysname}'s capacity to identify incidents from multimodal inputs, organize them into a hierarchy, and establish semantic connections between sensory data and incidents using a knowledge graph.  We also study the latency of processing multimodal information and storing it on the knowledge graph, and discuss the design decisions in our system leading aiming to lead to more efficient use of LLMs.  As a case study, we examine the 2025 Los Angeles Wildfires and showcase some of the relationships identified by different incidents and their associated multimodal information.




\section{Related Work}

\subsection{Multimodal Knowledge Graphs}

Existing works on multimodal knowledge graphs typically involve both semi-structured text and complex unstructured data such as images.  \cite{kannan_multimodal_2020} focuses on extracting text, images, and source code from academic publications and creating their respective graphs.  They apply a separate processing pipeline for each modality and attempt to align different entities into the Computer Science Ontology (CSO) \cite{noauthor_cso_nodate}.  \cite{wang_tiva-kg_2023} takes an existing knowledge graph (ConceptNet \cite{speer2017conceptnet}) and annotates it with additional attributes for images, sound, and video.  \cite{usmani_towards_2023} describes an ontology for structuring knowledge collected by sensors in smart cities.  However, it lacks a concrete implementation and doesn't collect event information necessary for smart city applications.

\subsection{Multimodal Data Integration}

The concept of unifying data sources which are diverse in type into a coherent view for monitoring or prediction has also been explored.  In the web domain, a \textbf{sensor web} \cite{broring2011new} describes an infrastructure that enables discovery and integration of sensors and their data.  An example of such a system is Senseweb \cite{grosky2007senseweb}, which is a web platform that ingests data from different sensory data streams and arranges them into a map view.  Some tools, such as Graph of Things \cite{le2016graph} describe an idea similar to ours, where they aim to construct a live knowledge graph of various IoT and social media data, and creating connections between heterogenous data sources using several ontologies.  While these platforms and tools share a similar goal to ours, they lack the ability to produce inferences and correlations between information which has historically required human intervention to directly process such correlations or create flexible pipelines to represent such reasoning.  It's also worth mentioning that while multimodal data fusion also brings together data from different sources, its objective is often task-specific and typically doesn't use human knowledge as explicitly as data integration might (such as the use of an ontology or linking against an external knowledge graph).  Thus we won't be including related work from this field.

\subsection{LLMs for Knowledge Graphs}

There are variety of works which utilize Large Language Models (LLMs) in various roles for interacting with knowledge graphs.  
\textbf{LLMs for constructing knowledge graphs}

\cite{holur_creating_2024} iteratively creates a semantic map of events occurring across news documents, and uses the LLM to fill out actors, relationships, and states given the raw text document.  \cite{giglou_llms4ol_2023} attempts to create an ontology from a document of terms, using some internal knowledge of the world to decide taxonomic and non-taxonomic relations.  \cite{zhu_llms_2024} provides knowledge graphs as answers to users questions by searching for relevant information and organizing it into facts represented as triples.

\textbf{LLMs expanding knowledge graphs with new facts and verifying existing facts}

\cite{xu_generate--graph_2024} uses an LLM in a Retrieval-Augmented-Generation (RAG) style to query a knowledge graph and answer a user query. In particular, it converts a natural language question into the appropriate query language, and passes the response as part of the context into the LLM for question answering.  \cite{toro_dynamic_2024} uses additional text from online sources to fill out empty fields in some given knowledge templates.  \cite{zaitoun_can_2023} uses an existing ontology and seeks to extend it with new relationships between entities given a collection of raw documents.

\cite{adam_traceable_2024} uses ground truth documents in order to verify triples found in knowledge graphs using LLMs.  \cite{boylan_kgvalidator_2024} works with several additional sources of obtaining evidence: web searches, LLM internal knowledge, and reference knowledge graphs.  

\textbf{LLMs for querying Knowledge Graphs}

\cite{edge_local_2025} stores nodes from a knowledge graph as document embeddings which are used at query time for RAG.  \cite{deng_graphvis_nodate} is given a graph as an input, either as a textual description or an image, along with a user query.  \cite{wu_retrieve-rewrite-answer_2023} aims to improve Q/A performance by rewriting extracted triples from a KG into more descriptive, readable text.  \cite{sui_can_2025} uses both a knowledge graph as well as some degree of the LLM's internal reasoning to answer more open ended Q/A questions. \cite{feng_knowledge_2023,sun_think_graph_2024,jiang_structgpt_2023,jin_graph_2024} allows for querying the knowledge graph by iteratively traversing the KG. These works treat the LLM as a planner for obtaining the necessary information to answer a question.

\subsection{Ontologies for Multimodal Sensory Knowledge Graphs}

Several ontologies were found to be relevant to our scenarios, and are particularly useful when discussing Semantic Web related platforms.  The Brick schema \cite{noauthor_brick_nodate} describes an RDF-written ontology, designed to represent resources in building management systems.  It includes concepts covering individual sensory devices as well as locations within buildings, represented by logical relationships. The Smart Applications Reference Ontology (SAREF) \cite{noauthor_saref_nodate} and its wearable version (SAREF4WEAR) \cite{noauthor_saref4wear_nodate} both describe sensor-specific characteristics, including properties of the device, measurements, and real world features to be measured.  The Sensor, Observation, Sample, Actuator Ontology (SOSA) \cite{noauthor_semantic_nodate} provides a similar sensor-centric view, describing features of interest, properties of sensors, as well as measurements. 

We found that these existing ontologies often lacked certain features necessary to organize urban sensory data (such as spatial, temporal, or incident classes).  Thus, we did not adopt any existing ontology as-is.  However, we have aligned a significant set of concepts with existing ontology classes and terms, which we describe later in Section \ref{background}
.  

While there exists some works on smart city modeling, such as \cite{noauthor_azureopendigitaltwins-smartcities_2025, espinoza-arias_ontological_2019, noauthor_smart_nodate}, these works either do not involve the same notion of events that we wish to model, lack sensor-specificity, or target simpler structured sensor readings and not more complex information such as actors within news articles.

\section{Background} \label{background}

In this section we provide an overview of terminology used in the {\sysname} system, as well as the ontology underlying the knowledge graph.

\subsection{Terminology}

Table \ref{tab:terms} describes the different terms we use when organizing data in urban spaces.  The notion of \textbf{incidents} is motivated by the need for larger occurrences influencing cities or states (such as natural disasters, major entertainment events, etc) to be represented in our knowledge graph, allowing us to identify more specific components contributing to them.  \textbf{Reports} are an abstraction where sensory data (and its associated metadata like time, accuracy, and location), as well as human-generated knowledge (such as news and tweets) can be integrated in our knowledge graph.  In our system, a report is an individual news article, image, or sample from a time series.  \textbf{Observers} can be used to identify the entity publishing a particular report.  For sensors, this can be a specific sensor ID, while for human-generated knowledge this can be an organization name (such as a news station) or username.  \textbf{Modality} represents a specific modality of a \textbf{report}.  This allows us to separate out different types of data present in a report possessing different properties.  For example, a news report may use images, audio, and text.  Images yield a caption property describing the image, audio describes the loudness of a scene, and text reveals the named actors involved an event.  \textbf{Aggregators} are the platforms from which we obtain data from (such as X.com, Caltrans CCTV, etc), and may vary in how trustworthy its knowledge is depending on the nature of its observers (though currently we do not model these attributes).  \textbf{Events} allow us to describe interactions between two parties, such as "accuse of a crime", or "protest" (where interactions are described under CAMEO codebook \cite{noauthor_data_nodate}).  In our current implementation, \textbf{events} are specific to news articles, but other modalities may also reveal specific interactions between parties.  \textbf{Actors} are the parties involved with an \textbf{event}, where both individual people and larger organizations of people may be named as actors.


\begin{table*}[]
\begin{tabular}{|l|l|}
\hline
Term        & Definition                                                                                                                                                                                                                                                                                                                                                                                                                                                                                                                  \\ \hline
Incident    & \begin{tabular}[c]{@{}l@{}}An occurrence, natural or manmade, that necessitates a response to protect life or property.  The word “incident” \\ includes planned events as well as emergencies and/or disasters of all kinds and sizes.  This definition is taken from\\ the Federal Emergency Management Agency (FEMA) \cite{noauthor_glossary_nodate}.  In our system, incidents may be linked to \\ multiple \textbf{reports}, and may occur as part of other incidents.\end{tabular} \\ \hline
Report      & \begin{tabular}[c]{@{}l@{}}An observation which may be structured or annotated in some way, produced from \textbf{observers}.  Annotations\\ involve labels such as timestamps, locations, or other labels which allow us to link the data to real world entities\end{tabular}                                                                                                                                                                                                                             \\ \hline
Observer    & \begin{tabular}[c]{@{}l@{}}An entity, either a sensor or individual that directly perceives, records and/or detects features of the environment.\\ These organize information into \textbf{reports} which are then collected by \textbf{Aggregators}.\end{tabular}                                                                                                                                                                                                                        \\ \hline
Modality    & \begin{tabular}[c]{@{}l@{}}A subset of a report represented by a unique type of data, such as image, audio, or text. A report may have multiple\\ modalities of information present.\end{tabular}                                                                                                                                                                                                                                                                                                                           \\ \hline
Aggregator  & \begin{tabular}[c]{@{}l@{}}Is an entity that collects information from multiple \textbf{observers} and presents them in the form of \textbf{reports}.\\ These sources may be sensory or human, and may represent information in multiple modalities, such as textual, \\ visual, auditory, and tabular.\end{tabular}                                                                                                                                                                      \\ \hline
Data Source & \begin{tabular}[c]{@{}l@{}}We use this term interchangeably with \textbf{Aggregator}, but when discussing our knowledge graph we prefer aggregator as it aligns \\ better with the rest of our terminology.\end{tabular}                                                                                                                                                                                                                                                                                                                      \\ \hline
Event       & \begin{tabular}[c]{@{}l@{}}Describes interactions between actors, with various terms relating to violent conflict, political actions, or other \\ types of cooperative behaviors.  We use the Conflict and Mediation Event Observations (CAMEO) codebook \cite{noauthor_data_nodate} to describe\\  our events.  \textbf{Incidents} can be viewed as events that become significant in terms of people involved or damage caused, \\ making them of interest to a city or state.\end{tabular}                                                                    \\ \hline
Actor       & \begin{tabular}[c]{@{}l@{}}A real world entity, person, or organization which has a name.  We use the Conflict and Mediation Event Observations\\  (CAMEO) codebook \cite{noauthor_data_nodate} to describe our actors.  However, while CAMEO is typically focused on larger groups \\ (e.g. political parties, unions, armed groups), we allow for individual people to be identified as actors.\end{tabular}                                                                                                                                          \\ \hline
\end{tabular}
\caption{List of terms used in our system and their definitions}
\label{tab:terms}
\end{table*}

\subsection{The {\sysname} ontology}

\begin{figure}
\centering
\includegraphics[width=8.5cm]{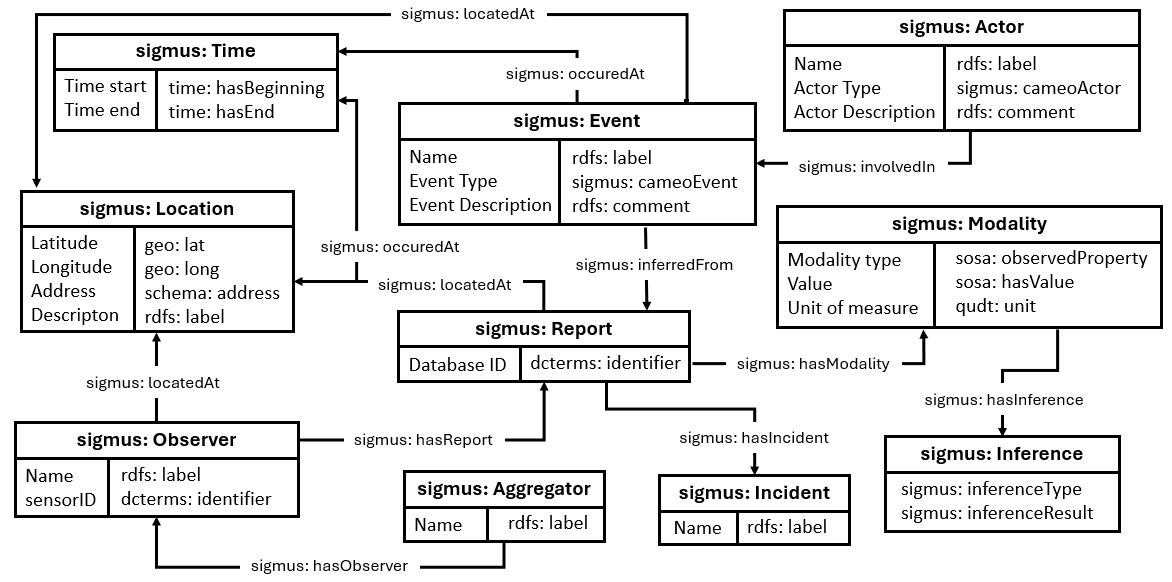}
\caption{The {\sysname} ontology}
\label{fig:ontology}
\vspace*{-15px}
\end{figure}

Having identified the main terms in our system, we now introduce the relationships between each of the terms, with each term being a class in our ontology).  Figure \ref{fig:ontology} describes how each class is related, as well as the main attributes present in for each class.

As our ontology focuses on creating common representations across different modalities (such as image, text, and tabular data), we designed classes with properties which can be shared across those data.  In addition, we also introduce new terms which allow us to represent high impact phenomena that influence cities and states, while also being able to attach different types of contributing factors and knowledge associated with such phenomena.  

Many of the properties are reused from existing ontologies:
\begin{itemize}
    \item \textit{dcterms: identifier} is used as a reference to a particular resource represented as some unique key or string, taken from the Dublin Core Terms ontology \cite{noauthor_dcmi_nodate}. 
    \item \textit{rdfs: label} and \textit{rdfs: comment} are used to describe a human readable string and a description, respectively.  These are taken from the RDF Schema \cite{noauthor_rdf_nodate}.
    \item \textit{time} properties are taken from the OWL Time ontology \cite{hobbs_ontology_2004}, allowing us to represent the start and end time of events.
    \item \textit{sosa} and \textit{qudt} are both part of the Semantic Sensor Network Ontology \cite{noauthor_semantic_nodate}, which we use to describe the properties and units of sensory measurements.  For more complex data (like images and audio), the \textbf{Value} field is treated as a filepath to the data rather than directly encoding it in the knowledge graph instance.
\end{itemize}

There are also several new fields:
\begin{itemize}
    \item The \textit{sigmus: Inference} class introduces the \textit{inferenceType} property, which describes the algorithm or ML task that was used to process the data (such as "Trend Analysis", "Image Captioning").  The result of this processing is stored in the \textit{inferenceResult} field as a human-readable string.
    \item The \textit{sigmus: cameoEvent} property represents the codes from CAMEO \cite{noauthor_data_nodate} that can be mapped to specific event terms and descriptions (e.g. assassinations) or actors (e.g. government).
\end{itemize}
\section{Methods} \label{methods}

\begin{figure*}
\includegraphics[width=18cm]{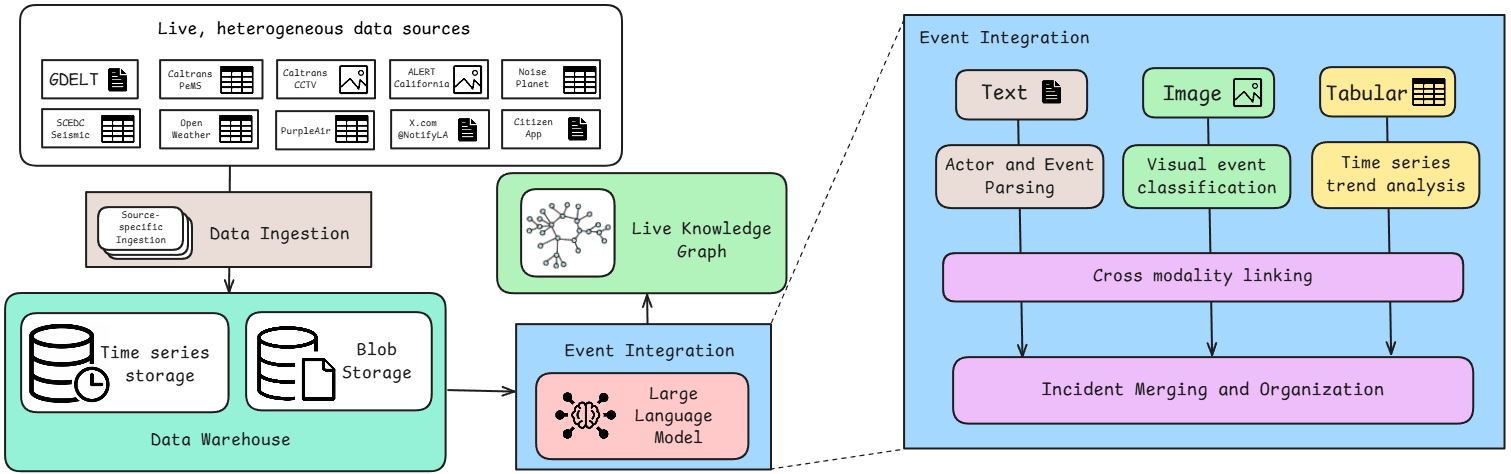}
\caption{Technical architecture for ingesting knowledge into a live knowledge graph for urban analysis.}
\label{fig:system}
\vspace{-1em}
\end{figure*}

\subsection{Data sources and Ingestion}

For our choices of data sources, had several criteria:
\begin{itemize}
    \item Publishes data on a regular basis such that important incidents can be monitored
    \item Capable of capturing incidents in a unique way, making it useful for integration
    \item Data can be obtained at low or no cost, reducing operational expenses of the {\sysname} system in the long term. 
\end{itemize}

To that end, we chose 10 different data sources, across 3 modalities: image, text, and tabular.  A summary of each data source is described in Table \ref{table:sources}.  

\begin{table*}[]
\begin{tabular}{|p{2cm}|p{4cm}|p{4cm}|p{1.5cm}|p{1cm}|p{3cm}|}
\hline
Data Source      & Attributes                                                    & Data Publish Method                      & In. \mbox{Frequency} & In. \mbox{Latency} & Data Filters                        \\ \hline
GDELT events \cite{noauthor_gdelt_nodate}     & news article URLs, named actors, CAMEO event codes, Geo names & Zip files of CSVs in 15 minute intervals & 15min    & 15min                     & Mentions of Los Angeles             \\ \hline
Caltrans PeMS \cite{california_pems_nodate}    & Traffic incidents, vehicle speeds, vehicle occupancy          & Zip files of CSVs updated daily          & 1 day           & 1 day                          & incidents and sensors in LA         \\ \hline
Caltrans CCTV \cite{noauthor_caltrans_nodate}   & images from highway cameras                                   & Live video streams                       & 15min    & None                           & Cameras in LA county                \\ \hline
ALERT California \cite{noauthor_alertcalifornia_nodate} & images from wildfire cameras                                  & Live video streams                       & 30min    & None                           & Cameras in LA county                \\ \hline
Noise Planet \cite{noauthor_noise-planet_nodate}     & noise levels in db(A)                                         & Zip files updated daily                  & 1 day           & 1 day                          & Measurements from LA                \\ \hline
SCEDC Seismic \cite{noauthor_southern_nodate}    & seismic waveform data from earthquake monitoring stations     & Zip files of waveform data updated daily & 1 day           & 1 day                          & Seismic stations in LA county       \\ \hline
OpenWeather \cite{noauthor_current_nodate}     & wind speed, precipitation, and weather descriptions           & API-based access, updated \~every 10 minutes                        & 1 hour          & None                           & Weather stations in LA county       \\ \hline
PurpleAir \cite{noauthor_real-time_nodate}        & PM2.5                                             & API-based access,  updated \~every 2 minutes                         & 1 hour          & None                           & Outdoor sensors in LA county        \\ \hline
X.com            & Tweet author, emergency type, location, tweet text            & Social network postings (event-based)                  & 15min    & 15min                     & LA emergency service accounts       \\ \hline
Citizen App \cite{noauthor_citizen_nodate}      & Text alerts, alert category, location of alert                & App alerts (event-based)                               & 15min          & 15min                          & Alerts about locations in LA county \\ \hline
\end{tabular}

\caption{Summary of data extracted from different data sources}
\label{table:sources}
\end{table*}

During the \textbf{Data Ingestion} step shown in Figure \ref{fig:system}, we have modality-specific processes for ingesting, filtering and organizing into our data warehouse.  This is necessary as each data source presents different methods of publishing their own data.  Our data warehouse provides both a common interface for interacting with all the ingested data, as well as durability for such data.  We implement the \textbf{Data Ingestion} for each modality using methods such as scraping, web API access, email alerts, or via a 3rd party capable of querying the data source at a lower cost (such as IFTTT for X.com).  After obtaining the data, we perform some degree of processing in order to obtain the described attributes in Table \ref{table:sources}, and filtering data involving Los Angeles County.  We choose Los Angeles County to demonstrate the {\sysname} system as it often has incidents occurring (such as sporting and cultural events, emergencies such as wildfires and earthquakes, and various political/social activities).   

While the \textbf{Attributes} column in Table \ref{table:sources} lists most of the modality-specific information our system ingests, it doesn't list metadata such as location, direction, time which is also used in our system.  The \textbf{In. Frequency} column describes the frequency at which we poll the data source in order to obtain new data.  Note that even for event-based sources (such as tweets or alerts), we interface with them via a 3rd party (such as an email server) which stores the data for us to poll.  The \textbf{In. Latency} column describes the worst case latency between the time data is published to the time our system ingests the data.  In some cases this is marked as \textit{none}, since the polling process obtains the most recent published measurement.

Our data warehouse stores the ingested data for durability before integrating different sources together into the live knowledge graph.  This data warehouse uses both time series storage (to allow for specialized to time-indexed data) while the blob storage allows for better scaling of unstructured data.

\subsection{Knowledge Graph and Integration}

The live knowledge graph acts as a working memory for integrating different data sources together and providing a visual interface for users to identify events and their connections.  To create this knowledge graph, we first must integrate our different data sources together and create connections based on our ontology.  

Each general modality type (image, text, and tabular) have different associated processes with them.

\begin{figure}
\includegraphics[width=9cm]{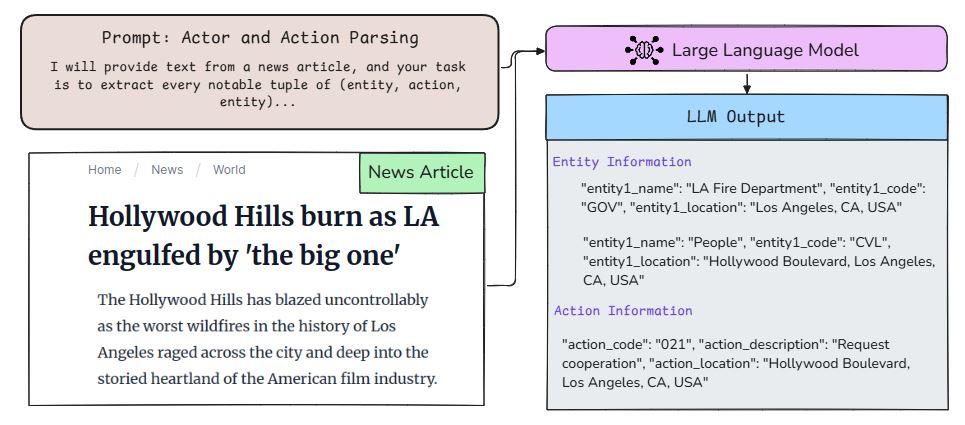}
\caption{Example of actor and event parsing}
\label{fig:textparse}
\vspace{-1em}
\end{figure}

\textbf{Actor and Event Parsing} uses LLMs to identify any real world organizations and persons involved in the text, as well as classifying these into the corresponding CAMEO term \cite{noauthor_data_nodate}.  In addition, it must categorize the event occurring between the actors using CAMEO event terms \cite{noauthor_data_nodate}, as well as identifying any alerts or emergencies using the Common Alerting Protocol (CAP) \cite{noauthor_common_nodate}, if any.  An high level example of this process is shown in Figure \ref{fig:textparse}.

In addition to establishing the actors and events, we also attempt to identify any noteworthy incidents.  As major incidents will have an associated name (i.e. Palisades Fire 2025) created by news agencies, we identify incidents primarily using the news modality (GDELT).  Later on, we use the incidents as a way of connecting different modalities together.

This process of identifying actors can be fairly time intensive, particularly as the LLM is prompted with a list of possible names which can be referring to the same actor (as well as the context, such as location, association with other actors, or other relevant news information).  As the number of actors present in our knowledge graph increases, this process can become incredibly inefficient especially as the number of input tokens increase.  Other problems may also arise as the input begins to reach the maximum input length of the LLM model.  Thus, we utilize a modified word edit distance which allows for relaxed penalties when matching part of names.  Using this, we obtain some top-k similar set of names which are used to query the model.

\begin{figure}
\includegraphics[width=9cm]{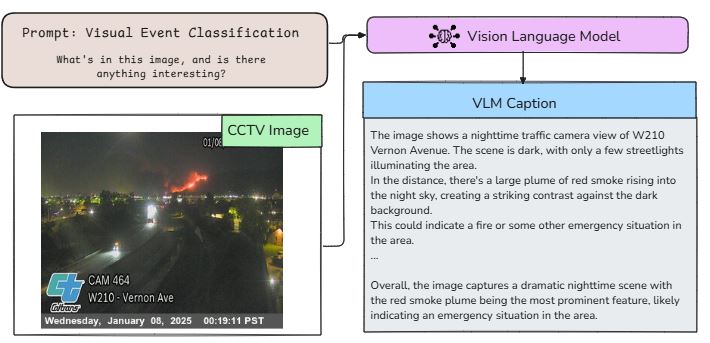}
\caption{Example of visual event classification}
\label{fig:visualclass}
\vspace{-1em}
\end{figure}

\textbf{Visual Event Classification} uses a Vision-Language Model (VLM) to caption images and identify interesting occurrences, if any.  An example of this process is shown in Figure \ref{fig:visualclass}.  Our main objective is to transform visual data into a human readable format which can both be stored on the knowledge graph, as well as utilized in downstream LLM reasoning. 

\textbf{Time Series Analysis} simply involves statistical analysis of values as well as a historical comparison at different intervals (e.g. 1 hour, 1 day, 1 week).  The result of this analysis is also stored on the knowledge graph, and allows for downstream LLM reasoning to identify anomalies or correlate values across modalities.

Having produced additional annotations for text, image, and tabular data, we create reports (i.e. from an image, a news article, a sensory reading).  After generating our reports, we can begin to link information across modalities via incidents.

\subsubsection{Cross Modality Linking} 

We connect reports from different modalities by linking them to the same shared incident.  In this work, we determine connections on every incoming report for each data source (aside from GDELT events, as that data source is focused on generating incidents).  

\begin{figure}
\includegraphics[width=9cm]{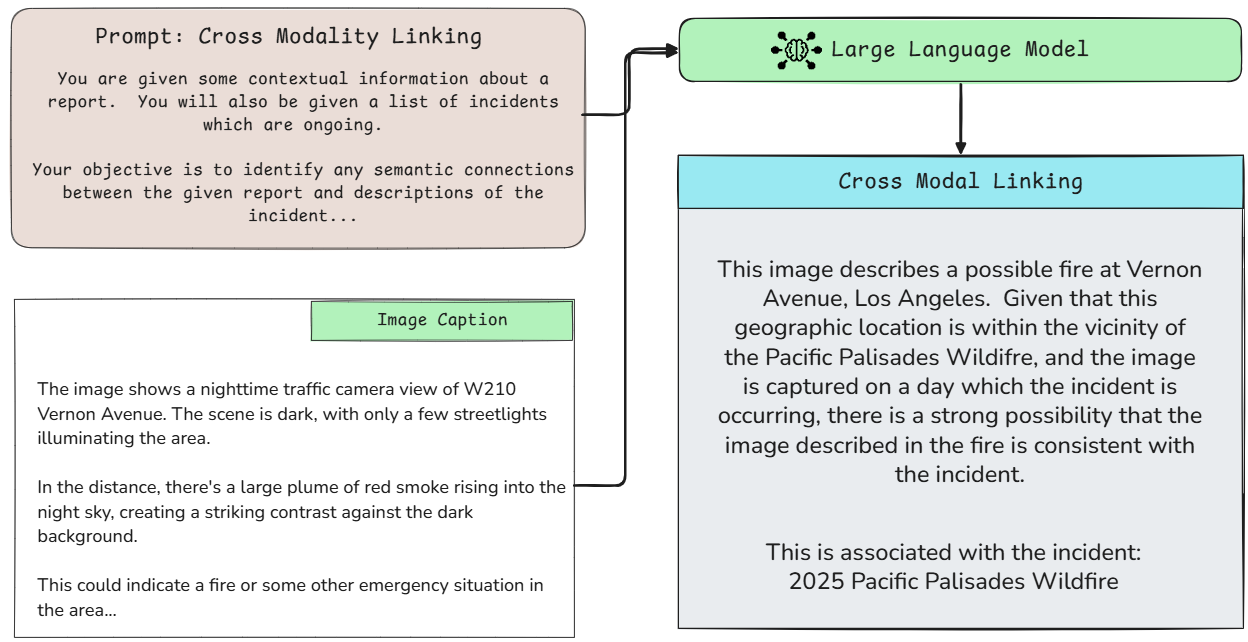}
\caption{Example of cross modal linking}
\label{fig:crossmodal}
\vspace{-1em}
\end{figure}

Once the report is processed depending on its modality (as described in the above sections), it is inserted into our knowledge graph.  After it is inserted into our knowledge graph, we query the knowledge graph for all incidents (as well as some additional related context, such as related time and geographic information for that incident).  We feed the given report's information (e.g. image caption, trends) into an LLM, along with the list of the incidents and contexts.  An example of this is shown in Figure \ref{fig:crossmodal}

\subsubsection{Incident Merging and Organization}

In order to both disambiguate different names which refer to the same incident, we use a RAG approach.  Similar to the problems arising in actor parsing, there may be many incidents, both ongoing and historical which may be relevant to a new incident.  In order to efficiently query the model, we also provide a similarity ranking of incidents outside of LLMs using a vector database and text embedding model to store embeddings of an incident and associated description of the incident.
 Given a new incident and its news text, we search through existing embeddings and rank them by similarity.  Finally, we use the top-k similar incidents and obtain the text of their original news article.  We feed this as context for the LLM to perform a selection over the top-k similar incidents to decide if the given incident can be organized under the same name.

\begin{figure*}
\includegraphics[width=18cm]{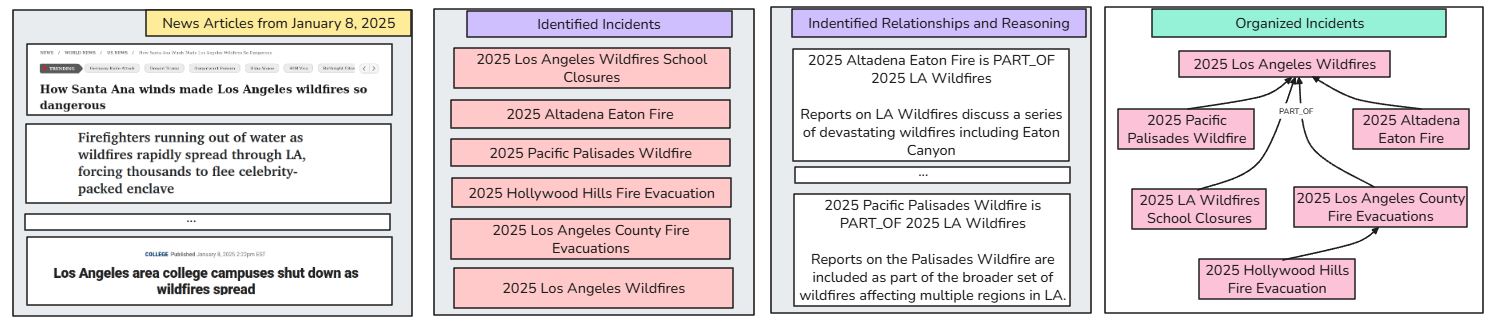}
\caption{Incident Merging and Organization}
\label{fig:incident_merge}
\vspace{-1em}
\end{figure*}



\section{Evaluation}

We evaluate {\sysname} using two LLMs.  One LLM is a DeepSeek-R1 70B model, which we operate on an H100 GPU server hosted on our local premises.  Another is the cloud-based OpenAI GPT-4o model.  The reason for these two different models is mainly due to quality of the output, with the GPT-4o performing significantly better than the DeepSeek model on the cross-modailty linking task.  For that reason, we rely on the GPT-4o model for the cross-modality linking task, while for all other tasks we utilize DeepSeek-R1.  For Visual Event classification, we rely on a Visual Large Language Model called NVILA \cite{liu_nvila_2025}, which also runs on the local H100 server.  During the Incident Merging and Organization task, we utilize a vector database called Marqo \cite{noauthor_marqo_nodate}, which allows us to find similar incidents based on their textual embeddings.  Our data warehouse uses TimescaleDB for its time series storage, and SeaweedFS for the blob storage.  All ingested data is organized and visualized in Neo4j, the knowledge graph used in {\sysname}.  In our evaluation, data storage services (i.e. time series, blob, vector, KG) operate on the same desktop PC. 

It is also worth noting that we mainly use the 2025 Los Angeles Wildfires as our case study.  As not all modalities are relevant or have data from this period of time, we only focus on a subset of the data sources described in Table \ref{table:sources}.  More specifically, we use GDELT events, PeMS, CCTV, OpenWeather, and PurpleAir. For the remaining sources, we either lacked data from that period of time, or they are not as relevant to the incident itself (i.e. seismic information).  


\begin{figure*}
\includegraphics[width=18cm]{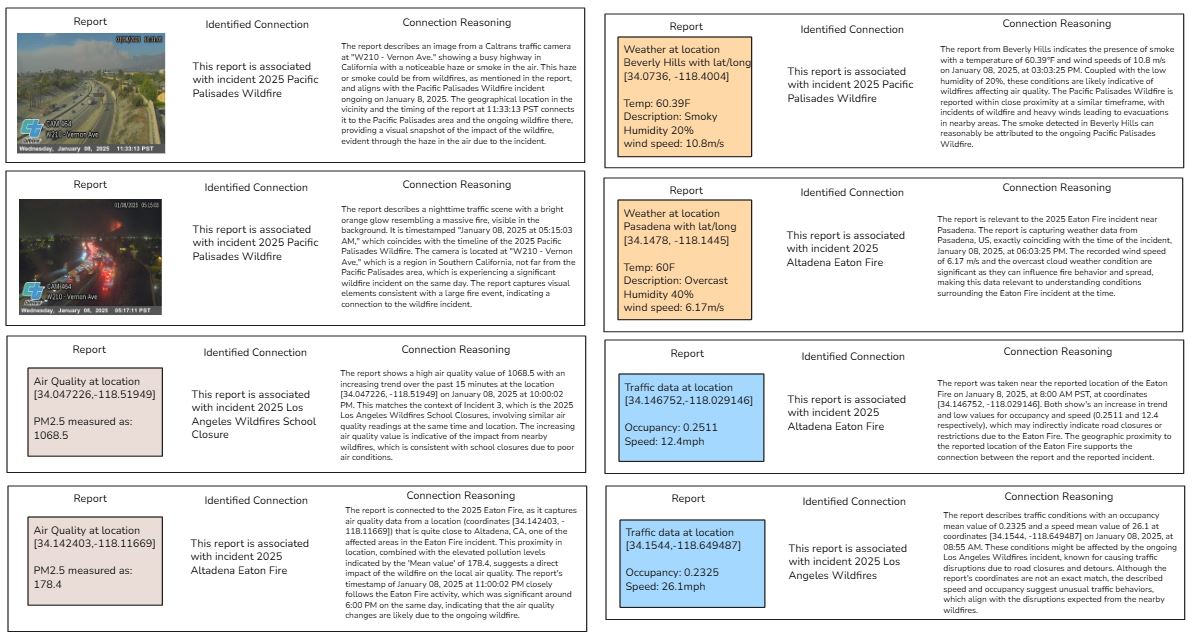}
\caption{Examples of linking multimodal reports with incidents}
\label{fig:report_info}
\vspace{-1em}
\end{figure*}





\begin{table}[]
\begin{tabular}{|l|c|c|}
\hline
Data Source      & \multicolumn{1}{l|}{Processing Latency (s)} & \multicolumn{1}{l|}{KG insert latency (s)} \\ \hline
GDELT events     & 39.3795                                     & 30.8118                                    \\ \hline
Caltrans PeMS    & 0.0105                                      & 0.0387                                     \\ \hline
Caltrans CCTV    & 5.905                                       & 0.0247                                     \\ \hline
ALERT California & 6.4577                                      & 0.0507                                     \\ \hline
Air Quality      & 0.0043                                      & 0.0207                                     \\ \hline
Weather          & 0.0144                                      & 0.0267                                     \\ \hline
\end{tabular}
\caption{Latency of ingesting a single report from each data source}
\label{table:latency}
\end{table}


Table \ref{table:latency} describes the average time, in seconds, to ingest a single report from each data source and add it to the knowledge graph.  While these times prove to be quite slow when performing historical analysis, the goal of {\sysname} is to execute and ingest data in a live manner - thus, these reported times are not batched operations.

As mentioned in section \ref{background}, a single report is treated as an individual news article, image, or row in a table.  Note that depending on the modality, the processing and knowledge graph latency may be affected.  For example, GDELT events (being a text modality) involves calling an LLM to identify actors, events, and incidents (which makes the bulk of the processing latency).  However, when inserting such data into a knowledge graph, we have to disambiguate names which refer to the same entity.  These are two kinds of disambiguation processes which occur when we insert items into the knowledge graph:  Actor disambiguation and Incident disambiguation.  Note that because event names are canonical (generated from \cite{noauthor_data_nodate}), we do not perform disambiguation for these.

\textbf{Actor Disambiguation} To disambiguate similar actors, {\sysname} first identifies a set of top-k similar actors based on word edit distance of their names, and prompts the LLM to identify whether these top-k actors (and their context, such as news report topic, location, etc) are likely to be the same actor.  On average, this process of actor merging takes 8.1101 seconds on average, with each report possibly involving multiple actors.  \\
\textbf{Incident Disambiguation} When inserting incidents into the knowledge graph, {sysname} identifies the top similar existing incidents based on both its label (e.g. "2025 Pacific Palisades Wildfire") as well as the original text from the article.  This process of ranking the top similar incidents is done via vector search in Marqo \cite{noauthor_marqo_nodate}, which produces embeddings of each incident and allows for similarity searching given a query.  After obtaining the top-k similar incidents, our system prompts the LLM to identify if the inserted incident already exists in the top-k.  In addition, the LLM also identifies if the inserted incident related to another incident, and we mainly focus on "IS\_PART\_OF" as a possible relationship between incidents.  This process of inserting incidents takes 7.8677 seconds on average.


\section{Discussion}

\subsection{Future Evaluation}
While much of our evaluation describes qualitative results for different modalities of data and some limited performance analysis of {\sysname}, there is still much evaluation to be done in the future.  First is determining a metric for evaluating the quality of connections made across modalities.  While for some cases this is fairly straightforward (e.g. CCTV image of a fire is captures information about a specific known fire incident) while for other cases it may be more complex (poor air quality at a particular location may be affected by a faraway incident such as smoke from a wildfire, or a local cooking action).  Verifying connectivity in these more complex cases requires significant user effort in examining both historical and neighborhood information to reach a conclusion.

There is also the issue of consistent outputs across different models.  Each LLM may produce a different set of connections within our knowledge graph, and it may be useful in future work to use voting strategies among a set of LLMs to determine the most likely set of connections.  A similar issue arises with the inputs to LLMs - small changes in the text prompt also affect performance, thus requiring some degree of prompt tuning to achieve the best performance.

\subsection{Case studies}

In this work we only examined the 2025 Los Angeles Wildfires.  While this incident itself is quite interesting (with multiple sub-incidents and relevance to multiple different data sources), our system may also be applied to other incidents.  In the future, we may consider major entertainment events (such as sports or music), or cases more relevant to domestic security, should they arise.  In addition, our geographic scope has thus far been limited to Los Angeles as our main urban space of interest, but other cities may present more interesting case studies as well.  As there is a plethora of public data sources collecting information for Los Angeles, we found it easier to start off in this location.

\section{Conclusion}

In this work, we introduced {\sysname}, a system that leverages Large Language Models (LLMs) to integrate and reason over multimodal urban data streams for the purpose of incident detection and knowledge organization. By unifying fragmented sensory inputs—including text, image, and tabular data—into a structured, semantically rich knowledge graph, {\sysname} enables a more comprehensive understanding of complex urban events such as wildfires, environmental hazards, or public safety incidents.

Through our case study of the 2025 Los Angeles wildfires, we demonstrated the system’s ability to qualitatively associate disparate data sources and derive contextually relevant relationships between them. Our findings suggest that LLMs can serve as powerful tools for bridging the gap between raw sensory input and actionable urban intelligence, particularly when world knowledge and contextual reasoning are required.

This work highlights the promise of LLM-powered knowledge systems in urban computing, especially for real-time monitoring and historical analysis of city-scale phenomena. Future directions include developing more case studies and incorporating more robust evaluation strategies for the quality and correctness of inferred relationships.


\begin{acks}
This research was sponsored in part by the Sandia National Laboratories award \#2169310 and the DEVCOM ARL award \#W911NF1720196.
\end{acks}

\bibliographystyle{plain}
\bibliography{Content/references.bib}










\end{document}